# BANGLA SIGN LANGUAGE RECOGNITION USING CONCATENATED BdSL NETWORK

Thasin Abedin
*Department of Electrical and Electronic Engineering*
*Islamic University of Technology (IUT)*
*thasinabedin@iut-dhaka.edu*

Ayana Moshruba
*Department of Electrical and Electronic Engineering*
*Islamic University of Technology (IUT)*
*ayanamoshruba@iut-dhaka.edu*

Khondokar S. S. Prottoy
*Department of Electrical and Electronic Engineering*
*Islamic University of Technology (IUT)*
*khondokarprottoy@iut-dhaka.edu*

Safayat Bin Hakim
*Department of Electrical and Electronic Engineering*
*Islamic University of Technology (IUT)*
*safayat.b.hakim@iut-dhaka.edu*

***Abstract*—Sign language is the only medium of communication for the hearing impaired and the deaf and dumb community. Communication with the general mass is thus always a challenge for this minority group. Especially in Bangla sign language (BdSL), there are 38 alphabets with some having nearly identical symbols. As a result, in BdSL recognition, the posture of hand is an important factor in addition to visual features extracted from traditional Convolutional Neural Network (CNN). In this paper, a novel architecture "Concatenated BdSL Network" is proposed which consists of a CNN based image network and a pose estimation network. While the image network gets the visual features, the relative positions of hand keypoints are taken by the pose estimation network to obtain the additional features to deal with the complexity of the BdSL symbols. A score of 91.51% was achieved by this novel approach in test set and the effectiveness of the additional pose estimation network is suggested by the experimental results**

## I. INTRODUCTION

Basically visual and manual modes are used in Sign Language in order to communicate. It is a rather complex mechanism specially for the general speaking and talking mass. So with the progress in the field of computer vision, much research has been done to detect SL. But the amount is rather small in case of detection of BdSl, [2], [6], [10], [13] being worth mentioning. There are 38 symbols of which 9 are vowels and 27 are consonants in BdSL. Machine Learning classifier, Support Vector Machine (SVM) was used by a lot of researchers to detect each alphabet from the hand gestures. But image classifiers, like CNN, are now popular among scientists to detect Sign Languages from images. The other methods are Principal Component analysis (PCA), skeleton detection, Hidden Markov Models etc [7], [9], [21]

In this paper a novel model, a combination of Image Network (CNN) [2], [6] and Pose Estimation Network is proposed by us.

A CNN model is used which was trained for visual feature extraction from the images. A pre-trained hand key estimation model, Openpose is used to estimate the hand key points from the image. The outputs from both the CNN and Openpose are then concatenated through 3 connected layers. The dataset used was the Bengali Sign Language Dataset obtained by the students of Bangladesh National Federation of the Deaf (BNFD).

## II. LITERATURE REVIEW

### A. Different Sign Language Detection

Of the 300 sign languages existing throughout the world, not many of them have sign language recognition tools . With the advancement in Artificial Intelligence, researchers have been interested to work on Sign language recognition to make life easier for the deaf-mute community. Among the other languages, most research has been performed on American Sign Language (ASL) detection.A real-time tool on the basis of desk and wearable computer based videos to recognise ASL was built by Starner et al. [16] and again Hidden Markov model was used in [15] which is also real-time applicable , In [20] kinetics was used for ASL recognition. Parallel Hidden Markov models was used in [18]. SL recognition systems have also been developing for other languages like Chinese Sign Language(CSL),Indian Sign Language(ISL), Japanese Sign Language(JSL), Arabic Sign Language, French Sign Language(FSL) etc. In [12] real time model was built for ISL. A classification technique based on Euclean distance which is Eigen value weighted was used by Singha et al in [14], For JSL recognition in [4] hand tracking system was built based on color, [17] showed hand feature extraction for JSL. For CSL recognition in [19] A phonemes based approach was taken. For CSL recognition a sign component based framework by using accelerometer and sEMG data in [8]. These are some of the notable works in Sign Language Recognition .

### B. Bangla Sign Language Detection

Bangla Sign language recognition is not yet much explored among the researchers. BdSL is actually quite unique. It is the modified form of American SL, British SL and Australian SL. Support Vector Machine (SVM) [1], K-Nearest Neighbour (KNN) and Artificial Neural Network (ANN) [11] were previously the common techniques used for BdSL recognition. But in the recent times CNN is the most popular among researchers for this job. In [13] Scale-Invariant Feature Transform (SIFT) was used to extract features and CNN for detection by SS Shanta et al. . VGG19 based CNN model for BdSL recognition was used by Abdul Muntakim Rafi and his team in [10]. In [6] CNN model was used by Md. Sanzidul Islam et al. to recognize BdSL digits. In [2] Faster R-CNN was used by Oishee Bintey Hoque and her team for real-time BdSL recognition. A novel BdSL recognition model combining both CNN and Pose estimation is introduced by our proposed model which is yet not done by other researchers before.

## III. Methodology

This paper proposes a novel architecture for BdSL recognition by combining an image network with a pose estimation network.

### A. Data Preprocessing

In this architecture, there are two separate inputs for image network and pose estimation network. For the image network input, the images are converted into grayscale. The images for pose estimation network input are in BGR format. Both the image inputs are normalized by the highest level (255) and resized into 64 x 64 for the task.

### B. Proposed Architecture

The proposed novel architecture consists of a separate image network and a pose estimation network. The two networks are tasked with separate but closely related purposes in terms of BdSL recognition. Our architecture builds a mechanism to efficiently use both the networks to give state of art results for Bangla sign language recognition. This paper refers to the proposed architecture as "Concatenated BdSL Network".

*1) Image Network:* The image network is tasked with the purpose of extracting visual features from pixel images. In this case for the image network, CNN was chosen for visual feature extraction as they provide better results compared to other machine learning approaches in case of feature extraction from images [10] For most of the works on BdSL recognition, pretrained CNN models were used due training difficulties, shortage of data etc. [13] However, the CNN model for this architecture is trained from the scratch. The efficiency of CNN is dependent on the architectural design. To select the best architecture, there is no universal rule and it is made by choosing the right number of convolution layers and neurons.

By hyper tuning various parameters, the best architecture was selected for this task. In this architecture, the CNN comprises of 10 convolution layers with batch normalization for each layer, 10 "ReLu" activation layers, 4 max pooling layers along with a single input and output layer. The convolution layers use trainable filters to detect the presence of specific features or patterns present in the input image to generate feature maps that are passed to the next layers of CNN. The "ReLu" activation layer proves to be the best choice for our image network as it converges faster than other activation functions by not activating all the neurons at the same time. It also does not saturate at the positive regions. After the activation layer, each layer is followed by a batch normalization. The max pooling layers further down sample the feature maps to highlight the most prominent features to be considered only. Finally, the output of the image network is flattened to a shape of 1x1x8196.

*2) Pose Estimation Network:* The task of recognizing the Bangla sign language is very complex as there are many letters in Bangla. Moreover, there is very subtle change among the gestures of some specific symbols that are even difficult to detect with human eyes. Hence, to aid our image network for this complex task of detecting the subtle differences among different hand gestures, a pose estimation network is included in this architecture. The main purpose of this pose estimation network is to estimate the hand keypoints from images. However, to train a pose estimation model from scratch is very computationally expensive and requires huge amount of data. For this reason, a pretrained hand keypoint estimation model from Openpose [10] is used. It estimates 21 co-ordinates points from a single hand image. So, our pose estimation network takes images as input and gives an output shape of 21x2. The final obtained output is flattened to a shape of 1x42 in a similar way to the image network output.

*3) Concatenated BdSL Network :* The features extracted from the image network and pose estimation network are complimentary to each other. As a result, combining both these features can greatly help to build a precise BdSL recognition model and tackle the complexities at the same time. To do so, the final flattened output from both the image and pose estimation networks are passed through two fully connected layers with "Relu" activation function and concatenated together. Afterwards, the new concatenated features are passed through 3 more fully connected layers, where the first two layers have "Elu" activation and the last one has "softmax" activation function. The proposed novel architecture in shown in details in Fig. 1.

The architecture merges the features extracted from the image network and pose estimation network.

### C. Training Method

Before the start of training, the two inputs are preprocessed accordingly as mentioned earlier. The preprocessed input of the image network is converted into a numpy file and fed directly into our proposed model. However, the preprocessed input for the pose estimation network has to be passed through the pretrained Openpose model at first. The obtained outputs are hand pose co-ordinate values that are converted into another numpy file and fed as the second input of our proposed model. The labels associated with the images are also converted into another numpy file. The same procedure is followed at the beginning of the evaluation, except for the labels.

It is also to be mentioned that during training time, as cost function, the cross entropy function is utilized. In order to reduce the cost function to a minimal value, a gradient descent based "Adam" optimization is used that has a learning rate of 0.001. The learning rate value is updated whenever the validation results of the model do not improve for 4 consecutive epochs using the reduce learning rate on plateau mechanism. For training, the weights are also initialized with small numbers. In order to avoid overfitting during training, early stopping mechanism is used to interrupt training after a certain time if we see that the validation loss has not improved upto a certain number of epochs.

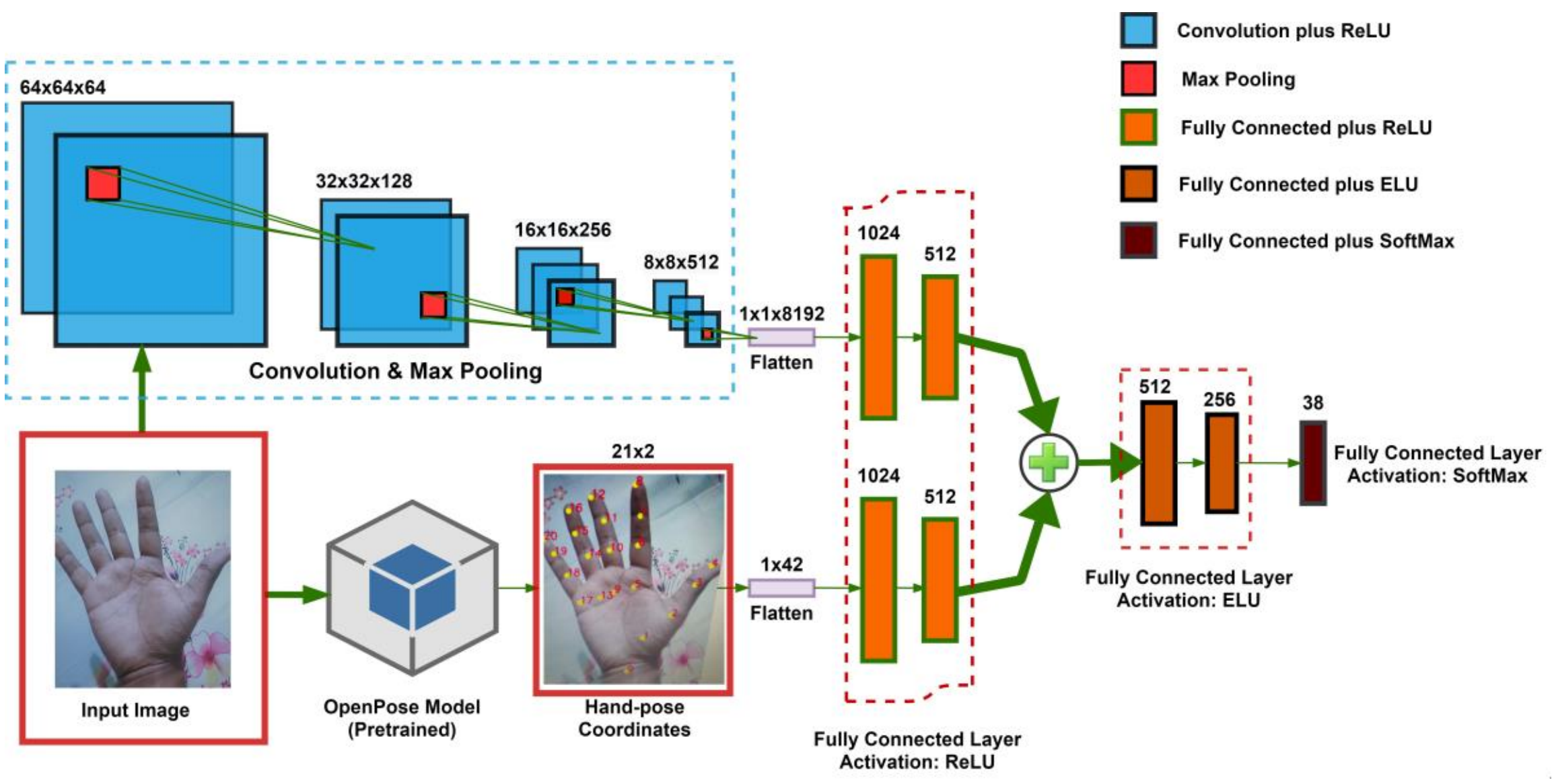

Fig. 1. Concatenated BdSL Network

## IV. Results

### A. Dataset And Experimental Setup

For this task, the Bengali Sign Language Dataset is used that was collected by Rafi et al. [10] from the students of Bangladesh National Federation of the Deaf (BNFD). It contains 11061 still images for training and 1520 for testing. All images are in their own 38 differently labeled folder. The dataset is splitted into training and validation set with 9959 and 1102 images respectively following previous works [10]. The experiments have been conducted under identical computational environment - Framework: *Tensorflow 2.2*, Platform: *Google Colaboratory having an 1-core allocated Intel Xeon processor with 2.2GHz and 12.72GB RAM*, GPU: *None*.

### B. Performance of Concatenated BdSL Network

Like mentioned earlier, most of existing BdSL recogniton models only extract features from CNN, which is termed as "Image Network" in this proposed architecture. In addition to this, the proposed "Concatenated BdSL Network" also extracts hand keypoint values from the images. As a result, to analyse the effectiveness of the additional features, a comparison is made between the proposed model with respect to existing CNN based model like modified VGG-19 Image Network [10] using the same dataset. Only the "Image Network" part of this proposed architecture has also been trained seperately and can be considered as another CNN based model. This model is referred to as "Only Image Network Model" in this paper and is used as the second model for comparison with the Concatenated BdSL Network. Both the "Only Image Network Model" and Concatenated BdSL Network has been trained for 30 epochs due to lack of computational resources like GPU.

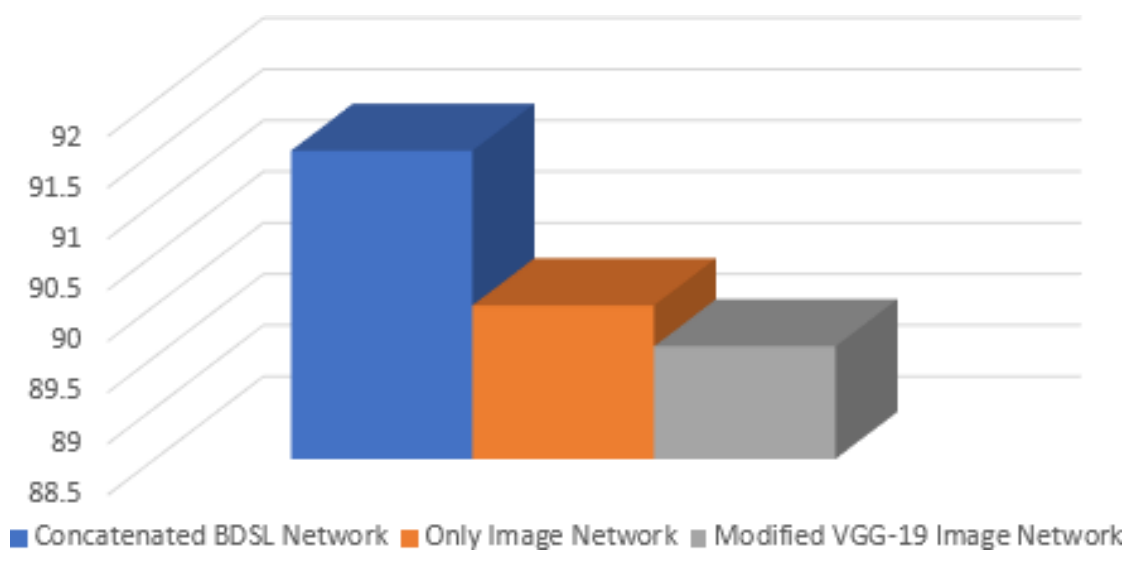

Fig. 2. Comparison of accuracy scores using test set among our different methods and modified VGG-19 Image Network

The proposed Concatenated BdSL Network architecture achieved validation score of 98.28 and overall test score of 91.51 whereas our "Only Image Network Model" got only 90 in test score. Another comparison with modified VGG-19 Image Network is given in Fig. 3. The confusion matrix of Concatenated BdSL Network is also included in Fig.5.

## V. Discussion

Some of the alphabet symbols in bangla sign language have very subtle difference with respect to others which makes the posture of hands very important for recognising BdSL alphabets. With this view in mind, one of the highlights of the proposed method is to extract additional hand keypoint features from the pose estimation network to deal with the difficulty of recognising Bangla sign language. By comparing with deeplearning based models that only extract visual features from CNN for Bangla sign language, the significance of hand pose estimation features is analysed. Even though the modified VGG-19 Image Network was trained on higher resolution images and for higher number of epochs, the Concatenated BdSL Network shows more promising result. It is also compared with "Only Image Network Model" which is trained on the Image Network part of the proposed architecture only, excluding the Pose Estimation Network. This is done to get a better understanding of the effectiveness of extracting hand keypoints.

|  | Concatenated BDSL Network | Modified VGG-19 Image Network |
|---|---|---|
| Training | **98.67** | 97.68 |
| Validation | **95.28** | 91.52 |
| Testing | **91.51** | 89.6 |
| Image size | **64*64** | 224*224 |
| Epoch | **30** | 100 |
| GPU | **No** | Yes |

Fig. 3. Comparison of classification accuracy and used resources between our novel method and other work with same dataset

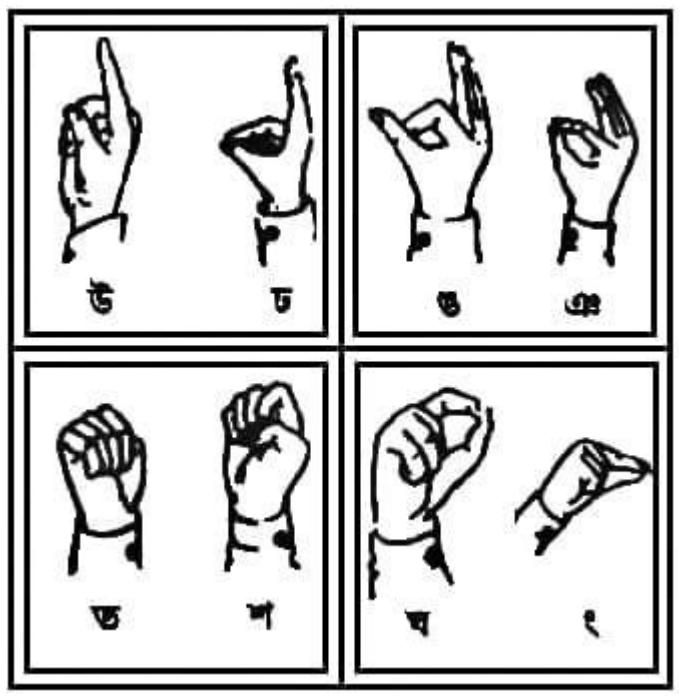

Fig. 4. Nearly identical signs of BdSL alphabet

The intuition is that the pose estimation network can estimate the relative position of a certain finger and its pose from the hand keypoint values because it assigns each finger joint of the hand with a unique id. This is an important factor in BdSL recognition to differentiate between two symbols that have very little difference. As a result, the Concatenated BdSL Network in addition to visual feature extraction from the image network, can also extract features to estimate finger positions at the same time that. So, combining both visual and positional information helps the novel Concatenated BdSL Network achieve higher precision. The proposed novel model achieved state of the art result on this Bengali Sign Language Dataset. However, further experiments could not be conducted with other similar works due to use of different dataset, like two handed bangla sign language [2], [5], very low training images [3] etc.

Despite being a very good classification model, the confusion matrix is not 100 percent diagonal as some misclassification is still prevalent. For general people even with human eyes, some signs are indistinguishable. They appear to be the same in terms of looks. The signs suffered mostly from recognizing these two symbols - [উ-ঢ]. From the confusion matrix that, it is also seen that the Concatenated BdSL predicted 'ঢ' instead of 'উ' 13 times out of 40 cases.

Other signs that suffered bad predictions are [ঙ-ঞ], [ত-শ], [ঘ-◌ং]. From Fig.4 it is seen that, these eight hand gesture are nearly identical to each other. So, there are still plenty of scope for improvement. If a pose estimation network could have been trained from scratch specifically for Bangla sign language hand pose estimation, these issues could have been further addressed. However, this requires computational resources and huge dataset that is beyond the scope of this work.

Collecting a Bangla sign language dataset, a pose estimation network can be trained from scratch in future work. With further work on the subject, the model can be developed to recognize real time continuous sign language and measures can be taken to make it suitable for commercial use.

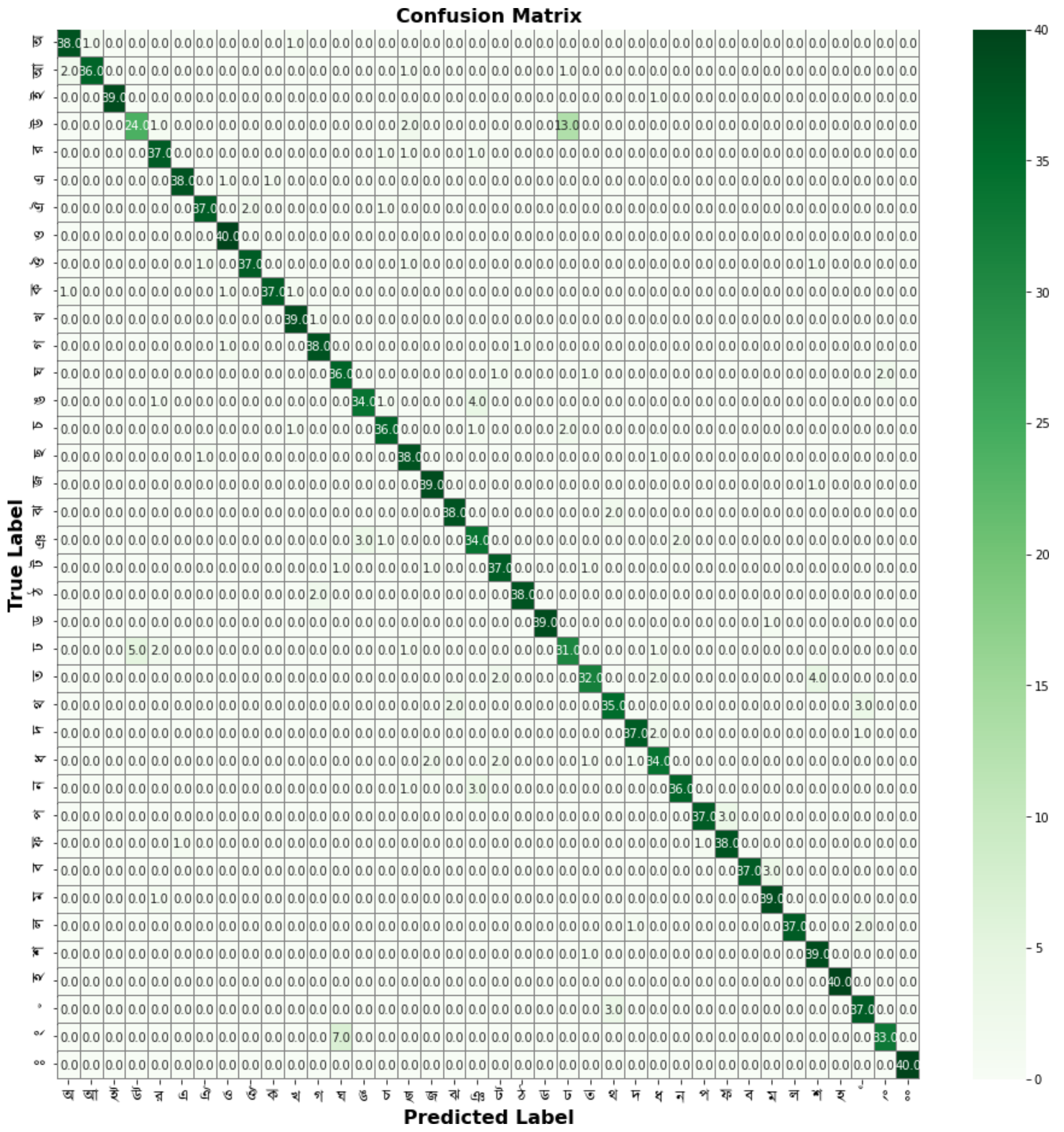

Fig. 5. Confusion Matrix of the BdSL alphabet